\pdfoutput=1
\def\year{2022}\relax
\documentclass[letterpaper]{article} 
\usepackage{aaai22}  
\usepackage{times}  
\usepackage{helvet}  
\usepackage{courier}  
\usepackage[hyphens]{url}  
\usepackage{graphicx} 
\usepackage{natbib}  
\usepackage{caption} 
\DeclareCaptionStyle{ruled}{labelfont=normalfont,labelsep=colon,strut=off} 
\usepackage{algorithm}
\usepackage[noend]{algpseudocode}

\usepackage{newfloat}
\usepackage{listings}
\floatstyle{ruled}
\newfloat{listing}{tb}{lst}{}
\floatname{listing}{Listing}

\usepackage{adjustbox}
\usepackage{amsmath}
\usepackage{amssymb}
\usepackage{booktabs}
\usepackage[symbol]{footmisc}
\usepackage[linguistics]{forest}
\usepackage{pgfplots}
\usepackage{siunitx}
\usepackage{subcaption}
\PassOptionsToPackage{dvipsnames,table}{xcolor}
\usepackage{tabularx}
\usepackage{tikz}
\usepackage{CJKutf8}

\def\corresponding{\footnote{Corresponding author.}}

\pgfplotsset{compat=newest,every axis/.append style={xticklabel style={font=\small},yticklabel style={font=\small},legend style={font=\small}}}
\usepgfplotslibrary{colorbrewer}

\usetikzlibrary{arrows.meta,backgrounds,calc,chains,matrix,positioning}
\tikzset{>=stealth}

\definecolor{cb-blue}{RGB}{0,119,187}
\definecolor{cb-green}{RGB}{0,153,136}
\definecolor{cb-red}{RGB}{204,51,17}
\definecolor{cb-yellow}{RGB}{153,119,0}

\algnewcommand{\LineComment}[1]{\State{\(\triangleright\) #1}}

\title{Hierarchical Context Tagging for Utterance Rewriting}
\author{
  Lisa Jin\textsuperscript{\rm 1}\thanks{Work done during an internship at Tencent AI Lab.},
  Linfeng Song\textsuperscript{\rm 2}\stepcounter{footnote}\corresponding{},
  Lifeng Jin\textsuperscript{\rm 2},
  Dong Yu\textsuperscript{\rm 2},
  Daniel Gildea\textsuperscript{\rm 1}
}
\affiliations{

  \textsuperscript{\rm 1}University of Rochester, Rochester, NY, USA\\
  \textsuperscript{\rm 2}Tencent AI Lab, Bellevue, WA, USA\\
  ljin14@cs.rochester.edu, \{lfsong, lifengjin, dyu\}@tencent.com, gildea@cs.rochester.edu
}

\newcommand{\cnr}{\textsc{CANARD}}
\newcommand{\mud}{\textsc{MuDoCo}}
\newcommand{\rew}{\textsc{Rewrite}}
\newcommand{\qua}{\textsc{QuAC}}
\newcommand{\bla}{\underline{\hspace{.75em}}}

\begin{document}

\maketitle

\begin{abstract}
  Utterance rewriting aims to recover coreferences and omitted information from the latest turn of a multi-turn dialogue. Recently, methods that tag rather than linearly generate sequences have proven stronger in both in- and out-of-domain rewriting settings. This is due to a tagger's smaller search space as it can only copy tokens from the dialogue context. However, these methods may suffer from low coverage when phrases that must be added to a source utterance cannot be covered by a single context span. This can occur in languages like English that introduce tokens such as prepositions into the rewrite for grammaticality. We propose a hierarchical context tagger (HCT) that mitigates this issue by predicting slotted rules (e.g., ``besides \bla{}'') whose slots are later filled with context spans. HCT (\textit{i}) tags the source string with token-level edit actions and slotted rules and (\textit{ii}) fills in the resulting rule slots with spans from the dialogue context. This rule tagging allows HCT to add out-of-context tokens and multiple spans at once; we further cluster the rules to truncate the long tail of the rule distribution. Experiments on several benchmarks show that HCT can outperform state-of-the-art rewriting systems by $\sim$2 BLEU points.
\end{abstract}

\section{Introduction}
Modeling dialogue between humans and machines has long been an important research direction with high commercial value. Such tasks include dialogue response planning \cite{li2016diversity}, question answering \cite{reddy2019coqa} and semantic parsing in conversational settings \cite{yu2019sparc}. Recent advances in deep learning and language model pre-training have greatly improved performance on many sentence-level tasks. However, machines are often challenged by coreference, anaphora, and ellipsis that are common in longer form conversations. Utterance rewriting \cite{kumar2017incomplete,su2019improving,elgohary2019can} has been proposed to resolve these references locally by editing dialogues turn-by-turn to include past context. This way, models need only focus on the last rewritten dialogue turn. Self-contained utterances also let models leverage sentence-level semantic parsers \cite{zhang2019amr} for dialogue understanding.

\begin{table}
  \small
  \centering
  \begin{tabular}{c l}
    \toprule
    Turn & Utterance\\
    \midrule
    $1$ & Why did Federer withdraw from the tournament?\\
    $2$ & He injured his back in yesterday's match.\\
    $3$ & Did he have any other injuries?\\
    \midrule
    $3^\ast$ & Did Federer have any other injuries besides his back?\\
    \bottomrule
  \end{tabular}
  \caption{Sample dialogue for utterance rewriting. Turns $1$--$2$ are the context, $3$ is the source, and $3^\ast$ is the target. The target is a context-independent rewrite of the source.}
  \label{fig:cnr-samp}
\end{table}

Much past work \cite{pan2019improving,elgohary2019can} on this task frames it as a standard sequence-to-sequence (seq-to-seq) problem, applying RNNs or Transformers. This approach requires re-predicting tokens shared between source and target utterances. To ease the redundancy, models may include a copy mechanism \cite{gu2016incorporating,see2017get} that supports copying source tokens besides drawing from a separate vocabulary. Yet generating all target tokens from scratch remains a burden. Perhaps as a result, these early efforts do not generalize well between data domains. For instance, \citet{hao2021robust} report a drop of 28 BLEU after transfer between two Chinese rewriting datasets \cite{su2019improving,pan2019improving}.

To exploit overlaps between source and target utterances, later work converts rewrite generation into source editing through sequence tagging. This tagging vastly simplifies the learning problem: predict a few fixed-length tag sequences, each with a small vocabulary. \citet{hao2021robust} propose a system that predicts edit actions to (\textit{i}) \textit{keep} or \textit{delete} a source token and (\textit{ii}) optionally \textit{add} a context span before the token. They rewrite Chinese datasets, where most targets can be covered by adding at most one context span per source token. Unfortunately, their single-span tagger is too brittle to insert out-of-context tokens or a series of non-contiguous spans, leading to low target phrase coverage.

Instead of separating edit action tagging from span insertion, \citet{liu2020incomplete} predict a word-level edit matrix between context-source pairs. In contrast to \citeauthor{hao2021robust}, this approach can add arbitrary non-contiguous context phrases before each source token. Though it may cover more target phrases, an edit matrix involves $\mathcal{O}(m)$ times more tags than a sequence for $m$ context tokens. Its flexibility also makes it easier to produce ungrammatical outputs, since any subset of context tokens can be added to the source. Finally, \citet{huang2021sarg} combine a source sequence tagger with an LSTM-based decoder. However, reverting back to a seq-to-seq approach introduces the same large search space issue that sequence tagging was designed to avoid.

To predict added phrases, we would like to keep the small search space of a span predictor while extending it to (\textit{i}) non-contiguous context spans and (\textit{ii}) tokens missing from the context altogether. For (\textit{i}), we first build a multi-span tagger (MST) that can autoregressively predict several context spans per source token. We use a syntax-guided method to automatically extract multi-span labels per target phrase. We further propose a hierarchical context tagger (HCT) that predicts a slotted rule per added phrase before filling the slots with spans. The slotted rules are learnt from training data and address (\textit{ii}) since they can include out-of-context tokens (e.g., determiners and prepositions). By conditioning a multi-span predictor on a small set of slotted rules, HCT can achieve higher phrase coverage than MST\@. By first planning rules and then realizing their slots, HCT dramatically enhances the performance gains of MST\@.\footnote{Our code is available at \url{https://github.com/lisjin/hct}.}

MST reaches the best baseline performance for an open-domain query rewriting task and is competitive on simpler benchmarks in English and Chinese \cite{su2019improving}. HCT achieves state-of-the-art results two of the three benchmarks. It improves by 1.9 and 2.7 BLEU on \cnr{} \cite{elgohary2019can} and \mud{} \cite{tseng2021cread}. In terms of robustness, it outperforms strong domain adaptation baselines on \mud{}\@.

\section{Multi-Span Tagger}
Our initial model balances between high coverage of target strings and a small model search space. RaST \cite{hao2021robust} achieves the latter by allowing only one context span to be added before each source token. In contrast, RUN \cite{liu2020incomplete} sacrifices search space for higher coverage by letting any subset of the context string replace or precede a source span. We first propose a multi-span tagger (MST) that expands RaST's coverage and shrinks RUN's search space.

As seen in Figure~\ref{fig:mst}, MST is composed of an action tagger on a source sequence and a semi-autoregressive span predictor over context utterances. It takes two token sequences as inputs: source $x = (x_1, \ldots, x_n)$ and context $c = (c_1, \ldots, c_m)$. For each source token, the action tagger decides whether or not to keep it---deleted tokens can later be replaced with context spans from the span predictor. In parallel, the multi-span predictor generates a variable-length sequence of context spans to insert before each source token.

\subsubsection{Encoder} We adopt BERT \cite{devlin2019bert} as our encoder. The tokens from context utterances $c$ are concatenated with source $x$ and fed into the encoder:
\begin{align}
  \label{eq:mst-enc}
  \mathbf{E}_c; \mathbf{E}_x = \text{BERT}(c; x),
\end{align}
where $\mathbf{E}_c \in \mathbb{R}^{m \times d}$ and $\mathbf{E}_x \in \mathbb{R}^{n \times d}$ are the resulting $d$-dimensional contextualized embeddings. This way, global information from $c$ and $x$ is encoded into both $\mathbf{E}_c$ and $\mathbf{E}_x$.

\subsubsection{Action Tagger} MST tags source token $x_i$ with a \textit{keep} or \textit{delete} action by linearly projecting its embedding $\mathbf{e}_i \in \mathbb{R}^d$, the $i$th row of $\mathbf{E}_x$:
\begin{align}
  \label{eq:mst-act}
  p(a_i \mid x_i) = \mathrm{Softmax}(\mathbf{W}_a\mathbf{e}_i),
\end{align}
where $\mathbf{W}_a \in \mathbb{R}^{2 \times d}$ is a learned parameter matrix.

\begin{figure}[t]
  \subcaptionbox{MST\label{fig:mst}}{
    \centering
    \begin{tikzpicture}
      \renewcommand{\sfdefault}{cmss}
      \begin{scope}[every node/.style={inner sep=.3em,align=center},
	gra/.style={fill=gray!25,rounded corners=2pt,font=\footnotesize},
	rnd/.style={draw,semithick,rounded corners=2pt,font={\scriptsize\sffamily}},
	cir/.style={draw,semithick,circle,inner sep=.1em,font={\scriptsize\sffamily}}]
	\node [rnd,text width=5em] (x0) {BERT};
	\node [gra,below of=x0,yshift=.5em] (x1) {$c\smash{;}\,x$};
	\node [gra,above left of=x0,yshift=.5em] (x3) {$\mathbf{E}\smash{_x}$};
	\node [gra,above right of=x0,yshift=.5em] (x4) {$\mathbf{E}\smash{_c}$};
	\node [rnd,above of=x3,anchor=east,draw=cb-yellow,fill=cb-yellow!15] (x5) {Action Tagger};
	\node [rnd,above of=x3,anchor=west,draw=cb-red,fill=cb-red!15] (x6) {Span Predictor};
	\node [cir,left of=x0,anchor=east,xshift=-1em] (x8) {1};
	\node [cir,above of=x3,yshift=1.5em] (x9) {2};
      \end{scope}
      \coordinate[above=.5em of x0.north] (aux);
      \draw [->] (x0) -- (aux) -| (x3);
      \draw [->] (aux) -| (x4);
      \path [->] (x1.north) edge (x0.south);
      \path [->] (x3.north) edge (x5.south);
      \path [->] (x3.north) edge (x6.south);
      \path [->] (x4.north) edge (x6.south);
    \end{tikzpicture}
  }
  \hfill
  \subcaptionbox{HCT\label{fig:hct}}{
    \centering
    \begin{tikzpicture}
      \renewcommand{\sfdefault}{cmss}
      \begin{scope}[every node/.style={inner sep=.3em,align=center},
	gra/.style={fill=gray!25,rounded corners=2pt,font=\footnotesize},
	rnd/.style={draw,semithick,rounded corners=2pt,font={\scriptsize\sffamily}},
	cir/.style={draw,semithick,circle,inner sep=.1em,font={\scriptsize\sffamily}}]
	\node [rnd,text width=5em] (x0) {BERT};
	\node [gra,below of=x0,yshift=.5em] (x1) {$c\smash{;}\,x$};
	\node [gra,above left of=x0,yshift=.5em] (x3) {$\mathbf{E}\smash{_x}$};
	\node [gra,above right of=x0,yshift=.5em] (x4) {$\mathbf{E}\smash{_c}$};
	\node [rnd,above of=x3,anchor=east,draw=cb-yellow,fill=cb-yellow!15] (x5) {Action Tagger};
	\node [rnd,above of=x3,anchor=west,draw=cb-blue,fill=cb-blue!15] (x6) {Rule Tagger};
	\node [rnd,above=4em of x4,xshift=.7em,draw=cb-red,fill=cb-red!15] (x7) {Span Predictor};
	\node [cir,left of=x0,anchor=east,xshift=-1em] (x8) {1};
	\node [cir,above of=x3,yshift=1.5em] (x9) {2};
	\node [cir,below of=x7,xshift=.8em,yshift=1.45em] (x10) {3};
      \end{scope}
      \coordinate[above=.5em of x0.north] (aux);
      \draw [->] (x0) -- (aux) -| (x3);
      \draw [->] (aux) -| (x4);
      \path [->] (x1.north) edge (x0.south);
      \path [->] (x3.north) edge (x5.south);
      \path [->] (x3.north) edge (x6.south);
      \path [->,bend right=2.25em] (x3.north) edge ($(x7.south east) - (0.2em, 0em)$);
      \path [->] (x6) edge (x7.south);
      \path [->] (x4.north) edge ($(x7.south east) - (0.2em, 0em)$);
    \end{tikzpicture}
  }
  \caption{Architecture comparison.}
  \label{fig:arch-cmp}
\end{figure}
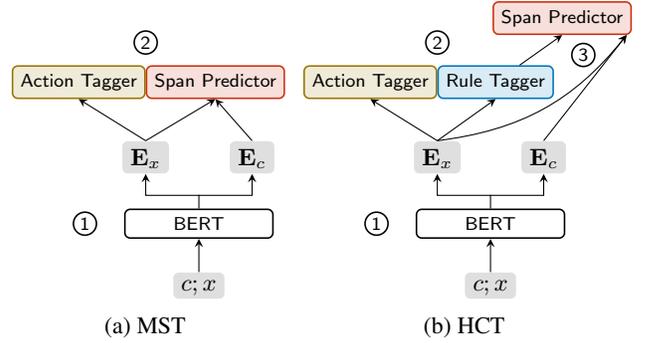

\subsubsection{Span Predictor} The span predictor outputs at most $l$ spans $\{s_{ij}\}_{j \leq l}$ from context $c$ to insert before each source token $x_i$. It predicts these spans autoregressively; the $j$th span $s_{ij}$ depends on all previous spans $\{s_{ij^\prime}\}_{j^\prime < j}$ at position $i$:
\begin{align}
  \label{eq:mst-sp}
  p(s_{ij} \mid c, x_i, j) = \text{MST}_s(c, x_i, \{s_{ij^\prime}\}_{j^\prime < j}).
\end{align}
We model generation of span $s_{ij}$ as predicting its start and end indices in context $c$. These two indices are captured through separate distributions over positions of $c$, given token $x_i$. Specifically, we apply additive attention \cite{bahdanau2015neural} to let source embedding $\mathbf{e}_i$ attend to all context embedding rows of $\mathbf{E}_c$. For example, the $j$th start index at source position $i$ is predicted as
\begin{align}
  \label{eq:sp-sing}
  p(s_{ij}^\uparrow \mid c, x_i, j) = \text{Attn}^\uparrow(\mathbf{E}_c, \mathbf{u}_{ij}),
\end{align}
where the $\uparrow$ indicates the start index distribution. The end index ($\downarrow$) is analogous in form. The joint probability of all spans $\{s_{ij}\}_{j \leq l}$ at source index $i$, denoted by $s_i$, is
\begin{align}
  \label{eq:sp-full}
  p(s_i \mid c, x_i) &= \prod_{j = 1}^l p(s_{ij} \mid c, x_i),
\end{align}
where \[p(s_{ij} \mid c, x_i) = p(s_{ij}^\uparrow \mid c, x_i, j)p(s_{ij}^\downarrow \mid c, x_i, j).\]

Because $s_{ij}$ depends on past spans indexed by $j^\prime < j$, the span predictor is considered semi-autoregressive for each source index $i$. Span prediction continues until either $j = l$ or $s^\uparrow_{ij}$ is a stop symbol (i.e., $0$), which can be predicted at $j = 0$ for an empty span. A span index at step $j$ depends on the attention distribution over context tokens at step $j - 1$:
\begin{subequations}
\begin{align}
  \mathbf{u}_{ij} &= \mathrm{ReLU}\left(\mathbf{W}_u\left[\hat{\mathbf{u}}_{ij}; \mathbf{u}_{i(j - 1)}\right]\right),\label{eq:sp-attn1}\\
  \hat{\mathbf{u}}_{ij} &= \sum_{k \in [1, m]} \alpha_{i(j - 1)k} \cdot \mathbf{e}_k,\label{eq:sp-attn2}
\end{align}
\end{subequations}
where $\alpha_{i(j - 1)k}$ is the attention coefficient\footnote{Details can be found in Appendix \ref{sec:appendix}.} between $c_k$ (embedded by $\mathbf{e}_k$) and $x_i$ and $\mathbf{W}_u \in \mathbb{R}^{d \times 2d}$. Similar to the notion of coverage in machine translation \cite{tu2016modeling}, this helps maintain awareness of past attention distributions.

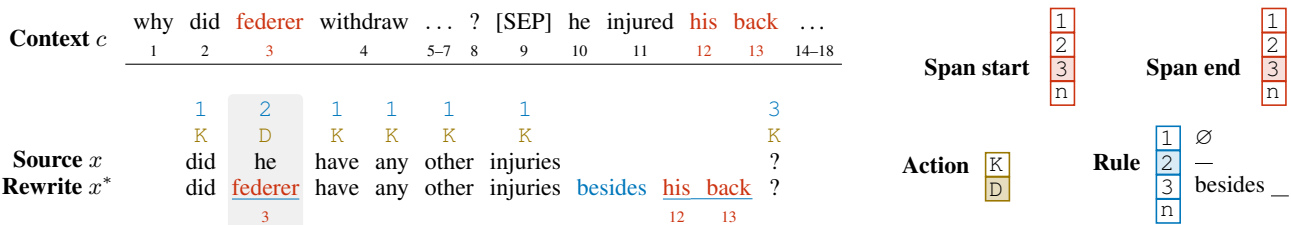
\begin{figure*}[t]
  \centering
  \begin{tikzpicture}[baseline=(x1.north)]
    \begin{scope}[every node/.style={font=\footnotesize,inner sep=0.1em}],
      \node (x1) {
	\addtolength{\tabcolsep}{-3pt}
	\begin{tabular}{*{12}c}
	  why & did & \color{cb-red}federer & withdraw & $\ldots$ & ? & [SEP] & he & injured & \color{cb-red}his & \color{cb-red}back & \ldots\\
	  \tiny{1} & \tiny{2} & \color{cb-red}\tiny{3} & \tiny{4} & \tiny{5--7} & \tiny{8} & \tiny{9} & \tiny{10} & \tiny{11} & \color{cb-red}\tiny{12} & \color{cb-red}\tiny{13} & \tiny{14--18}\\
	  \hline
	\end{tabular}
	\addtolength{\tabcolsep}{3pt}
      };
      \node [left=0.25em of x1] (x0) {\textbf{Context} $c$};
      \node [below=4.7em of x0] (x4) {\textbf{Rewrite} $x^\ast$};
      \node [below=3.7em of x0] (x3) {\textbf{Source} $x$};
      \node [below=1em of x1,anchor=north] (x2) {
	\addtolength{\tabcolsep}{-3pt}
	\begin{tabular}{*{10}c}
	  \color{cb-blue}\texttt{1} & \color{cb-blue}\texttt{2} & \color{cb-blue}\texttt{1} & \color{cb-blue}\texttt{1} & \color{cb-blue}\texttt{1} & \color{cb-blue}\texttt{1} & & & & \color{cb-blue}\texttt{3}\\
	  \color{cb-yellow}\texttt{K} & \color{cb-yellow}\texttt{D} & \color{cb-yellow}\texttt{K} & \color{cb-yellow}\texttt{K} & \color{cb-yellow}\texttt{K} & \color{cb-yellow}\texttt{K} & & & & \color{cb-yellow}\texttt{K}\\
	  did & he & have & any & other & injuries & & & & ?\\
	  did & \color{cb-blue}\underline{\color{cb-red}federer} & have & any & other & injuries & \color{cb-blue}besides & \color{cb-blue}\underline{\color{cb-red}his}{\rlap{\color{cb-blue}\underline{\hspace{2.6em}}}} & \color{cb-red}back &  ?\\
	      & \color{cb-red}\tiny{3} & & & & & & \color{cb-red}\tiny{12} & \color{cb-red}\tiny{13}\\
	\end{tabular}
	\addtolength{\tabcolsep}{3pt}
      };
    \end{scope}
    \begin{scope}[on background layer]
      \draw [draw=none,fill=gray!12,rounded corners=2pt] (-3.4,-2.55) rectangle (-2.4,-.75);
    \end{scope}
  \end{tikzpicture}
  \hspace{1em}
  \begin{tikzpicture}[baseline=(c1.north),start chain=1 going below,start chain=2 going below,start chain=3 going below,start chain=4 going below,node distance=-.04em]
    \begin{scope}[every node/.style={semithick,inner sep=.45em},
      lbl/.style={font={\small}}]
      \foreach \x in {1,2} {
	\pgfmathsetmacro{\coly}{ifthenelse(\x==2,"cb-yellow!25","none")}
	\node [draw=cb-yellow,fill=\coly,on chain=1] (a\x) {};
      }
      \node [lbl,left=.1em of a1] {\textbf{Action}};
      \node [lbl,right of=a1] {\texttt{K}};
      \node [lbl,right of=a2] {\texttt{D}};
      \foreach \x in {3,4} {
	\node [on chain=1] (a\x) {};
      }
      \foreach \x in {1,2,3,4} {
	\pgfmathsetmacro{\colb}{ifthenelse(\x==2,"cb-blue!15","none")}
	\pgfmathsetmacro{\colr}{ifthenelse(\x==3,"cb-red!15","none")}
	\pgfmathsetmacro{\ncust}{ifthenelse(\x<4,\x,"n")}
	\node [draw=cb-blue,fill=\colb,on chain=2,right=5.5em of a\x,yshift=1em] (b\x) {};
	\node [lbl,right of=b\x] {\texttt{\ncust}};
	\node [draw=cb-red,fill=\colr,on chain=3,above=3.5em of b\x,xshift=-4em] (c\x) {};
	\node [lbl,right of=c\x] {\texttt{\ncust}};
	\node [draw=cb-red,fill=\colr,on chain=4,above=3.5em of b\x,xshift=4em] (d\x) {};
	\node [lbl,right of=d\x] {\texttt{\ncust}};
      }
      \node [lbl,left=0.1em of b2] {\textbf{Rule}};
      \node [lbl,right=0.05em of b1] {$\varnothing$};
      \node [lbl,right=0.05em of b2] {\bla{}};
      \node [lbl,right=0.05em of b3] {besides \bla{}};
      \node [lbl,left=0.1 of c3] {\textbf{Span start}};
      \node [lbl,left=0.1 of d3] {\textbf{Span end}};
    \end{scope}
  \end{tikzpicture}
  \caption{HCT sample run. After encoding, every source token $x_i$ is tagged with an action $a_i$ and rule $r_i$. Context span(s) for $x_i$ are then predicted based on $r_i$ and cross-attention between source and context embeddings. For instance, $x_3 = \text{``he''}$ is tagged with $a_2 = \mathrm{\textit{delete}}$ and rule $r_2 = \mathrm{\bla{}}$. The slot in $r_2$ is filled with the $c$ span $s_2 = (3,3)$ that contains ``federer''.}
  \label{fig:arch}
\end{figure*}

\subsubsection{Optimization} MST is trained to minimize cross-entropy over gold actions $a$ and spans $s$:
\begin{align}
  \label{eq:xc-loss}
  \mathcal{L}_e &= -\sum_{i = 1}^n\log p(a_i \mid x_i)p(s_i \mid c, x_i).
\end{align}
Since MST runs in parallel over source tokens, output sequences may be disjointed. MST optimizes sentence-level BLEU under an RL objective \cite{chen2014systematic} to encourage more fluent predictions. Besides minimizing cross-entropy in Eq.~\ref{eq:xc-loss}, MST maximizes similarity between gold $x^\ast$ and sampled $\hat{x}$ as reward term $w$:
\begin{align}
  \label{eq:rl-loss}
  \mathcal{L}_r &= -\Delta(\hat{x}, x^\ast)\log p(\hat{x} \mid c, x) = -w\log p(\hat{x} \mid c, x)
\end{align}
where $\Delta$ denotes sentence-level BLEU\@. In practice, we use the greedily decoded $\hat{x}_g$ as a baseline reward relative to $\hat{x}$ to reduce variance in REINFORCE:
\begin{align*}
  w &= \Delta(\hat{x}, x^\ast) - \Delta(\hat{x}_g, x^\ast).
\end{align*}
This encourages the model to generate both accurate and fluent sentences. To further stabilize gradient updates, we follow \citet{kiegeland2021revisiting} by scaling the rewards by the minimum ($\min$) and maximum ($\max$) $r$ values in the current batch: $r = (r - \min) / (\max - \min)$. This provides batch-level normalization across rewards.

The final loss is a weighted average over cross-entropy and RL losses in Eqs.~\ref{eq:xc-loss} and \ref{eq:rl-loss}:
\begin{align}
  \label{eq:fin-loss}
  \mathcal{L} = (1 - \lambda)\mathcal{L}_e + \lambda\mathcal{L}_r,
\end{align}
where $\lambda$ is a scalar weight empirically set to $0.5$.

\section{Hierarchical Context Tagger}
While MST supports more flexible context span insertion, it can not recover tokens that are missing from the context (e.g., prepositions). We propose a hierarchical context tagger (HCT) that uses automatically extracted rules to fill this gap.

As shown in Figure~\ref{fig:hct}, HCT adopts the BERT encoder (Eq.~\ref{eq:mst-enc}) and action tagger (Eq.~\ref{eq:mst-act}) from MST\@. In addition, HCT includes a rule tagger that chooses which (possibly empty) slotted rule to insert before each source token. HCT is composed of two levels: both action and rule taggers run in parallel at the first level, then the second level span predictor takes tagged rules as inputs. Since the span predictor fills in a known number of slots per rule, it no longer needs to produce stop symbols as it does in MST (Eq.~\ref{eq:mst-sp})\@.

\subsubsection{Rule Tagger} The rule tagger linearly projects the embedding of source token $x_i$ to choose a rule to insert before it:
\begin{align}
  p(r_i \mid x_i) = \mathrm{Softmax}(\mathbf{W}_r\mathbf{e}_i),
\end{align}
where $\mathbf{W}_r$ parameterizes a rule classifier of $p$ rules that includes the null rule $\varnothing$ for an empty insertion. The rule set is automatically extracted from training data.

\subsubsection{Span Predictor} The second-level span predictor expands rule $r_i$ containing $k \ge 1$ slots into spans $s_i = (s_{i1}, \ldots, s_{ik})$:
\begin{align}
    p(s_{ij} \mid c, x_i, r_i, j) &= \text{HCT}_2(c, x_i, r_i, \{s_{ij^\prime}\}_{j^\prime < j}),
\end{align}
where $1 \leq j \leq k$. Unlike MST, HCT learns rule-specific slot embeddings to anchor each span to a rule $r_i$. Instead of conditioning spans $s_i$ on all tokens $x$ and rules $r$, we find it sufficent to restrict it to a single $x_i$ and $r_i$.

To condition its span predictor on tagged rules, HCT learns contextualized rule embeddings using the same input token BERT encoder. Slots at the same relative position across rules are represented by the same special slot token. For example, the rule ``\bla{} and \bla{}'' is assigned the tokens (\texttt{[SL0]}, \texttt{and}, \texttt{[SL1]}), whereas the rule ``\bla{}'' is simply \texttt{[SL0]}. Embeddings of these \texttt{[SL\textasteriskcentered{}]} tokens are learned from scratch and allow relative positional information to be shared across rules. A special \texttt{[CLS]} token is prepended to a rule's token sequence before applying the BERT encoder, and its embedding is used to represent the rule.

We bias context-source attention (Eq.~\ref{eq:sp-sing}) on a rule embedding by updating the query embedding $\mathbf{e}_i$ as
\begin{align*}
  \mathbf{e}_i = \mathrm{ReLU}\left(\mathbf{W}_c\left[\mathbf{e}_i; \mathbf{r}_i\right]\right),
\end{align*}
where $\mathbf{W}_c \in \mathbb{R}^{d \times 2d}$ is a learned projection matrix. Eq.~\ref{eq:sp-sing} can then be replaced by
\begin{align}
  p(s_{ij}^\uparrow \mid c, x_i, r_i, j) = \text{Attn}^\uparrow\left(\mathbf{E}_c; \mathbf{u}_{ij}\right).
\end{align}

We note that HCT's nested phrase predictor can also be seen as learning a grammar over inserted rules. Each source token is preceded by a start symbol that can be expanded into some slotted rule. Rules come from a fixed vocabulary and take the form of a sequence of terminal tokens and/or slots (e.g., ``\bla{} by \bla{}'' or ``in \bla{}''). In contrast, slots are nonterminals that can only be rewritten as terminals from the context utterances (i.e., spans). While slotted rules are produced from start symbols in a roughly context-free way---conditioned on the original source tokens---terminal spans within a rule are not. Spans in the same rule are predicted autoregressively to support coherency of successive spans.

\subsubsection{Optimization} Following Eq.~\ref{eq:xc-loss}, HCT's loss is:
\begin{align}
  \label{eq:xc-loss2}
  \mathcal{L}_e &= -\sum_{i = 1}^n\log p(a_i \mid x_i)p(r_i \mid x_i)p(s_i \mid c, x_i, r_i),
\end{align}
where $p(s_i \mid c, x_i, r_i)$ is analogous to Eq.~\ref{eq:sp-full}. HCT optimizes the same RL objective as MST by replacing $p(\hat{x} \mid c, x)$ in Eq.~\ref{eq:rl-loss} with $p(\hat{x} \mid c, x, r)$. Its total loss is the same as Eq.~\ref{eq:fin-loss}.

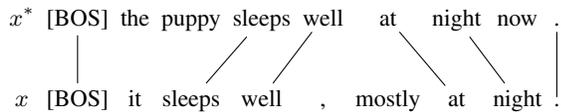
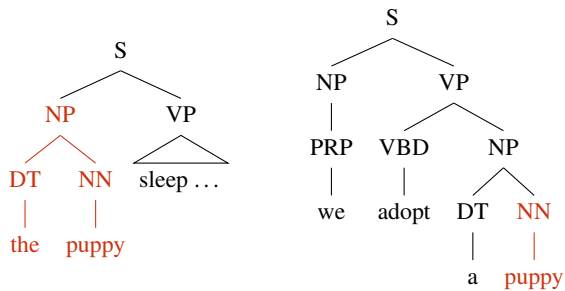
\begin{figure}[t]
  \begin{subfigure}{0.47\textwidth}
    \small
    \centering
    \begin{tikzpicture}
      \begin{scope}[every node/.style={font=\footnotesize,inner sep=.25em}]
	\matrix[column sep=,row sep=2em,matrix of nodes] {
	  \node (x0) {$x^\ast$}; & \node (x1) {[BOS]}; & \node (x2) {the}; & \node (x3) {puppy}; & \node (x4) {sleeps}; & \node (x5) {well}; & \node (x6) {at}; & \node (x7) {night}; & \node (x8) {now}; & \node (x9) {.};\\
	    \node (y0) {$x$}; & \node (y1) {[BOS]}; & \node (y2) {it}; & \node (y3) {sleeps}; & \node (y4) {well}; & \node (y5) {,}; & \node (y6) {mostly}; & \node (y7) {at}; & \node (y8) {night}; & \node (y9) {.};\\
	  };
      \end{scope}
      \draw (x1) -- (y1);
      \draw (x5) -- (y4);
      \draw (x4) -- (y3);
      \draw (x6) -- (y7);
      \draw (x7) -- (y8);
      \draw (x9) -- (y9);
    \end{tikzpicture}
    \caption{Token-aligned sentences to illustrate LCS usage.}
    \label{fig:lcs}
  \end{subfigure}
  \begin{subfigure}{0.47\textwidth}
    \forestset{sn edges/.style={for tree={parent anchor=south, child anchor=north,align=center,base=bottom,inner sep=.2em}},
    redst/.style={for tree={text=cb-red},for descendants={edge=cb-red}}}
    \begin{minipage}{.43\textwidth}
      \centering
      {\small\begin{forest}
	sn edges [S [NP,redst [DT [the]] [NN [puppy]]] [VP [sleep \ldots,roof]]]
      \end{forest}}
    \end{minipage}
    \hfill
    \begin{minipage}{.55\textwidth}
      \centering
      {\small\begin{forest}
	sn edges [S [NP [PRP [we]]] [VP [VBD [adopt]] [NP [DT [a]] [NN,redst [puppy]]]]]
      \end{forest}}
    \end{minipage}
    \caption{Subtree of missing target (left) and that of the context (right). Note that both strings are lemmatized.}
    \label{fig:syn-tree}
  \end{subfigure}
  \caption{Examples of automatic action and span labeling.}
  \vspace{-0.5em}
\end{figure}

\section{Automatic Label and Rule Extraction}
\label{sec:data-anno}
Since most datasets for utterance rewriting do not provide alignments between target phrases and context spans, we apply methods to extract them automatically. The LCS algorithm extracts actions and context-target alignments for span insertions. A bottom-up syntactic method attempts to align phrases that LCS failed to find context spans for. While these alignment methods are shared by MST and HCT, only HCT depends on rule extraction.

\subsection{LCS Action and Phrase Alignment}
\label{sec:lcs-align}
To annotate actions and phrases added to source $x$, we compute the longest common subsequence (LCS) between $x$ and target utterance $x^\ast = (x_1, \ldots, x_l)$. The LCS algorithm relies on dynamic programming and runs in time $\mathcal{O}(nl)$. We traverse token alignment pairs in LCS n-grams from left to right, extracting tags based their positions in $x^\ast$ and $x$. Figure~\ref{fig:lcs} shows an example alignment. Concretely, if the alignment pairs are
\begin{enumerate}
  \item adjacent in both sequences: \textit{keep} the aligned tokens (e.g., ``sleeps well'').
  \item only adjacent in $x$: \textit{keep} the aligned tokens (e.g., ``night .''), \textit{add} the phrase between pairs in $x^\ast$ (``now'').
  \item only adjacent in $x^\ast$: \textit{keep} the aligned tokens (e.g., ``well at'') and \textit{delete} tokens between pairs in $x$ (``, mostly'').
  \item adjacent in neither $x$ nor $x^\ast$: \textit{keep} the aligned tokens (e.g., ``[BOS] sleeps'') and \textit{delete} tokens between pairs in $x$ (``it''), \textit{add} those between pairs in $x^\ast$ (``the puppy'').
\end{enumerate}
This produces the desired actions $a$ and a list of unaligned target phrases that map to context spans. Phrases that can be found as single context spans are kept, while the remaining ones are searched for in the next step.

\subsection{Syntax-Guided Phrase Alignment}
\label{sec:syn-align}
Unaligned phrases are common in datasets like \cnr{}, where only $\sim$42\% of dialogue utterances are free from such phrases. We extract context spans that partially cover the phrases for additional model supervision during training. To do so, we lemmatize \cite{manning2014stanford} the unaligned phrase and the context, then try to align syntactic spans of the target phrase to context spans. For example, ``the puppy'' may be missing from the context but ``puppy'' aligns to a span (Figure~\ref{fig:syn-tree}, right). Each target span must perfectly match some context span, and we find the set of aligned spans with highest coverage of the unaligned target phrase. In HCT, the target tokens that remain unaligned to a context span can be covered by tokens in an extracted rule.

Searching only for syntactic spans lets us avoid na\"{i}vely exploring all possible splits of the target phrase. Span-level instead of purely token-level alignment (similar to RUN) also limits how many insertions the model must predict per phrase. Furthermore, we align the lemmatized target phrase and context strings to ignore inflection changes. This way, target tokens inflected differently from those in the context (e.g., ``sleeps'' and ``slept'') map to the same lemma (e.g., ``sleep'') and can still align.

\begin{algorithm}[t]
  \caption{Target alignment to context spans}
  \label{alg:syn-align}
  \begin{algorithmic}[1]
    \Require Target subtree $ttr$, context string $c$
    \Ensure Context spans $sps = [s_1, \ldots, s_c]$
    \Procedure{DescendTree}{$ttr, i, k, c$}
    \LineComment{Find $n$-length span $s^\prime$ of $c$ matching $ttr[i, k]$}
    \State $s^\prime, n \gets \Call{AlignSpan}{ttr, i, k, c}$
    \State $sps, n\_sum \gets [\,], 0$
    \ForAll{$(i2, k2) \in ttr.children$}
    \State $ch\_sps, n2 \gets \Call{DescendTree}{ttr, i2, k2, c}$
    \State $sps \gets sps \oplus ch\_sps$
    \State $n\_sum \gets n\_sum + n2$
    \EndFor
    \If{$n\_sum > n$} \Comment{Adopt best spans of children}
    \State \Return $sps, n\_sum$
    \EndIf
    \State \Return $[s^\prime], n$
    \EndProcedure
    \State $sps, n \gets \Call{DescendTree}{ttr, 0, |ttr.leaves|, c}$
  \end{algorithmic}
\end{algorithm}

For each missing phrase, we parse \cite{kitaev2019multilingual} the constituency tree of its target utterance $x^\ast$, then isolate the smallest subtree $ttr$ covering it (e.g., left subtree of Figure~\ref{fig:syn-tree}). Next, we iterate over subtree $ttr$ to find the context spans $sps$ that most overlap with the target phrase (Alg.~\ref{alg:syn-align}). We first try to match the current target constituent's span $[i, k]$ to a span in context $c$ (line 2), where $n$ is the character count of the matching spans' strings---$0$ if no match is found. We then find the best spans $ch\_sps$ covering the constituent's children (lines 3--7). If their summed match lengths $n\_sum$ is greater than $n$, then the children's spans replace the constituent's (lines 8--10). The result is a list of the best-matching context spans per unaligned target phrase.

\begin{table}[t]
  \setlength{\arraycolsep}{.6125em}
    \[
      \begin{array}{l *{7}c}
	\toprule
	& \text{B}_1 & \text{B}_2 & \text{B}_4 & \text{R}_1 & \text{R}_2 & \text{R}_\text{L}\\
	\midrule
	\text{Pro-Sub} & 60.4 & 55.3 & 47.4 & 73.1 & 63.7 & 73.9\\
	\text{Ptr-Gen} & 67.2 & 60.3 & 50.2 & 78.9 & 62.9 & 74.9\\
	\text{RUN} & \mathbf{70.5} & 61.2 & 49.1 & 79.1 & 61.2 & 74.7\\
	\text{RaST} & 53.5 & 47.6 & 38.1 & 62.7 & 50.5 & 61.9\\
	\text{MST} & 66.6 & 59.9 & 48.7 & 79.5 & 64.1 & 79.0\\
	\text{HCT} & 68.7 & \mathbf{62.3} & \mathbf{52.1} & \mathbf{80.0} & \mathbf{66.5} & \mathbf{79.4}\\
	\bottomrule
      \end{array}
    \]
    \vspace{-.5em}
    \caption{BLEU-$n$ ($\text{B}_n$) and ROUGE-$n/\text{L}$ ($\text{R}_{n/\text{L}}$) on \cnr{}. Pro-Sub, Ptr-Gen, and RUN results are drawn from their respective works.}
    \label{tab:cnr}
\end{table}

\subsection{Rule Extraction and Clustering}
\label{sec:rule-extract}
We extract rules from the single- or multi-span alignments from the previous two subsections, where each span maps to a rule slot. Recall that slotted rules can (\textit{i}) bind together multiple spans in the same target phrase and (\textit{ii}) include out-of-context target tokens. For (\textit{i}), rules consist entirely of slots (e.g., ``\bla{} \bla{}'' for $k = 2$), resembling glue rules introduced in machine translation \cite{chiang2005hierarchical}. Rules in (\textit{ii}) arise when a phrase maps to non-contiguous context spans; the uncovered target sub-phrases are left as-is in the rule (e.g., ``the'' in rule ``the \bla{}'').

Some rules extracted from the training set may be rare, existing in the long tail of the rule distribution. To concentrate the rule vocabulary on higher frequency rules, we cluster lexically similar rules and map low-frequency rules to high-frequency ones (e.g., ``in addition to \bla{}'' $\to$ ``in addition \bla{}''). Though this adds noise to low-frequency rules, we assert that the cost is outweighed by reducing the rule vocabulary size.

We apply affinity propagation \cite{frey2007clustering} clustering on the negative token-level LCS distance of rules. We use LCS rather than Levenshtein distance to disallow token substitutions that can unevenly affect short rules. A rule slot counts as a single token, and we normalize the LCS distance to fall between 0 and 1. Affinity propagation is ideal as it does not require tuning the number of clusters. Briefly, it iteratively clusters rules based on their suitability with exemplars, or representative points. The suitability of a point $i$ and exemplar $k$ depends both on other exemplars $k^\prime$ of the same point and other points $i^\prime$ of the same exemplar.

After clustering, we further filter the rule clusters based on the proportion of data points that each cluster covers. If a representative rule falls below a preset threshold (e.g., 0.5\% of the points), then all rules in the cluster are labeled with the glue rules that match their slot counts.

\section{Experiments}
We evaluate HCT on several multi-turn utterance rewriting datasets---two in English and one in Chinese.

\subsection{Setup}
\subsubsection{Datasets} \cnr{} \cite{elgohary2019can} is derived from \qua{} \cite{choi2018quac}, a question answering dialogue dataset about Wikipedia articles. The first two utterances are usually an article's title and section heading. Next is an exchange between a student, who asks questions to learn about the text, and a teacher, who answers questions by providing text excerpts.

\mud{} \cite{tseng2021cread} includes conversations between a system and user on multiple domains (e.g., calling or news). Notably, it is about half the size of \cnr{} and contains much shorter dialogues. Furthermore, the majority of its dialogues require no edits, in contrast to \cnr{}. Its relative simplicity may reflect the task-oriented rather than information-seeking nature of its dialogues.

\begin{table}[t]
\setlength{\arraycolsep}{.31em}
    \[
	\begin{array}{l *{8}c}
	\toprule
	& \multicolumn{2}{c}{\text{Calling}} & \multicolumn{2}{c}{\text{Messag.}} & \multicolumn{2}{c}{\text{Music}} & \multicolumn{2}{c}{\text{All}}\\
	\cmidrule(lr){2-3}\cmidrule(lr){4-5}\cmidrule(lr){6-7}\cmidrule(lr){8-9}
	& \text{B}_4 & \text{EM} & \text{B}_4 & \text{EM} & \text{B}_4 & \text{EM} & \text{B}_4 & \text{EM}\\
	\midrule
	\text{Joint} & 84.3 & \mathbf{77.7} & 83.1 & 68.8 & 61.8 & 40.9 & 80.4 & 69.3\\
	\text{RaST} & 83.3 & 75.2 & 82.0 & 69.1 & 66.1 & 44.6 & 80.2 & 68.5\\
	\text{MST} & 82.9 & 73.6 & 81.4 & 64.7 & 64.4 & 40.2 & 81.2 & 65.8\\
	\text{HCT} & \mathbf{86.3} & 76.3 & 84.0 & \mathbf{69.8} & \mathbf{71.8} & \mathbf{47.3} & 83.1 & \mathbf{69.7}\\
	\hspace{.75em}\text{-RL} & 86.1 & 75.9 & \mathbf{84.2} & 68.6 & 70.7 & 45.6 & \mathbf{84.1} & 69.2\\
	\bottomrule
      \end{array}
    \]
    \vspace{-.5em}
    \caption{BLEU-4 ($\text{B}_4$) and exact match accuracy (EM) on \mud{}. Only three of the six domains are shown. The ``-RL'' line ablates BLEU rewards under an RL objective.}
    \label{tab:mud}
  \end{table}

\rew{} \cite{su2019improving} contains Chinese conversational dialogues crawled from social media platforms. It is comparable in size to \mud{}, yet virtually all of its targets differ from the source utterances. Since Chinese lacks inflection and has fewer function words than English, the majority of phrases ($87.57\%$ of \rew{} dialogues) added to \rew{} sources can map to single context spans.

\subsubsection{Baselines} We compare against results reported in the following baseline systems. RaST \cite{hao2021robust} can be seen as a special case of MST, where the maximum number of rule slots $l = 1$. Pro-Sub is a simple baseline from \citet{elgohary2019can} that replaces the first pronoun in a source with the main entity of the dialogue (e.g., article title). Ptr-Gen is the LSTM-based hybrid pointer generator network of \citet{see2017get}. RUN is the edit matrix tagging approach by \citet{liu2020incomplete}. It has an LSTM-based encoder and applies a semantic segmentation network to tag the edit matrix. Joint is the coreference resolution and query rewriting model from \citet{tseng2021cread}.

\subsubsection{Evaluation} We measure utterance rewriting quality via widely used automatic metrics: BLEU, ROUGE, and EM\@. The BLEU \cite{papineni2002bleu} score between the predicted and gold target utterances is reported on all datasets. For \cnr{} and \rew{}, we add BLEU-$n$ and ROUGE-$n$ for $n \in \{1,2\}$ to evaluate smaller n-grams, in addition to ROUGE-L \cite{lin2002manual}. EM refers to string-level exact match accuracy and is the least forgiving.

\subsubsection{Implementation details} We use the uncased and Chinese variants of \texttt{BERT-base} \cite{wolf2020transformers} as the respective encoders for English and Chinese experiments. \textit{Adam} is used to optimize the full model with a learning rate of \num{1e-5} for \cnr{} and \mud{} and \num{2e-5} for \rew{}. We apply early stopping after 15 epochs once the development set BLEU-4 stops growing three epochs in a row.

\subsection{Results}
\subsubsection{\cnr{}} As shown in Table~\ref{tab:cnr}, HCT achieves the highest score on nearly all query rewriting metrics. It surpasses the previous top model, Ptr-Gen, by margins of 1.9 BLEU-4 and 4.5 ROUGE-L. By extension, it exceeds the edit matrix and sequence tagger baselines of RUN, RaST, and MST\@. We note that adding adding multi-span prediction between RaST and MST provides a sizeable performance boost: 10.6 BLEU-4 and 17.1 ROUGE-L\@. Adding hierarchy to MST via HCT's rule predictor gives a boost of 3.4 BLEU-4. The expressivity of multi-span tagging and structure of rule prediction appear to boost rewrite quality.

\begin{table}[t]
  \setlength{\arraycolsep}{.31em}
  \[
    \begin{array}{l *{8}c}
      \toprule
      & \multicolumn{2}{c}{\text{Calling}} & \multicolumn{2}{c}{\text{Messag.}} & \multicolumn{2}{c}{\text{Music}} & \multicolumn{2}{c}{\text{All}}\\
      \cmidrule(lr){2-3}\cmidrule(lr){4-5}\cmidrule(lr){6-7}\cmidrule(lr){8-9}
      & \text{B}_4 & \text{EM} & \text{B}_4 & \text{EM} & \text{B}_4 & \text{EM} & \text{B}_4 & \text{EM}\\
      \midrule
      \text{RaST} & 84.7 & 74.9 & 80.5 & 66.7 & 61.9 & 39.5 & 81.6 & 67.1\\
      \text{MST} & 84.1 & 74.0 & 79.6 & 65.0 & 62.6 & 39.5 & 81.7 & 66.0\\
      \text{HCT} & \mathbf{86.0} & \mathbf{75.9} & \mathbf{83.4} & \mathbf{69.1} & \mathbf{63.2} & \mathbf{40.9} & \mathbf{82.8} & \mathbf{68.4}\\
      \bottomrule
    \end{array}
  \]
  \caption{BLEU-4 ($\text{B}_4$) and EM on \mud{} for models trained on the calling domain only.}
  \label{tab:mud-dom}
\end{table}

\subsubsection{\mud{}} Table~\ref{tab:mud} summarizes rewriting results for multi-domain coreference utterances. Scores are grouped by the main domains---calling, messaging, and music---that respectively cover about 54\%, 20\%, and 14\% of the dataset. Overall, HCT outperforms the Joint baseline of \citet{tseng2021cread} by 2.7 BLEU-4\@. Within the smaller messaging and music domains, HCT scores 1 and 6.4 EM points higher than Joint. The trend holds to a lesser degree for the RaST baseline as well. Thus, HCT generalizes even better on lower-data domains than the simpler baselines of RaST and MST\@. This suggests that HCT's slotted rules improve data coverage without hurting model robustness across domains.

We also use \mud{} as a testbed for domain adaptation experiments. In particular, we focus on zero-shot generalization of models on unseen data domains. We train models on the calling domain only and evaluate on the full dataset that includes five unseen domains. Besides the three shown in Table~\ref{tab:mud-dom}, the full \mud{} dataset also contains news, reminders, and weather domains. Since RaST generalizes well on cross-domain Chinese utterance rewriting, it is useful for domain adaptation evaluation. HCT generalizes especially well on the messaging and music domains; compared to RaST, it gains 2.9 BLEU-4 and 2.4 EM points for the former and 1.3 BLEU-4 and 1.4 EM for the latter. On the full dataset, HCT achieves a 1.3 higher EM score than RaST\@. This is especially surprising given that HCT was trained on rules extracted only from the calling domain; its ability to generalize these rules to other domains is promising.

\subsubsection{\rew{}} As a final benchmark, we consider how well HCT can rewrite utterances in a different language like Chinese. The \rew{} dataset is a challenging setting for HCT since over 90\% of its dialogues can be covered using single-span insertions from the context; this is likely due to linguistic differences between Chinese and English. Thus, the better expressivity offered by a multi-span predictor only benefits a small proportion of the examples. However, Table~\ref{tab:rew} shows that HCT scores 1.5 BLEU-4 and 1.9 EM points above RaST\@. Compared to RUN, it achieves a 2.2 higher ROUGE-L score. RUN may score higher on metrics over small n-grams due to its addition of a fixed set of unseen tokens to each context, improving target token coverage.

\begin{table}[t]
  \setlength{\arraycolsep}{.45em}
  $$
  \begin{array}{l *{7}c}
    \toprule
    & \text{B}_1 & \text{B}_2 & \text{B}_4 & \text{R}_1 & \text{R}_2 & \text{R}_\text{L} & \text{EM}\\
    \midrule
    \text{RUN} & \mathbf{93.5} & \mathbf{90.9} & \mathbf{85.5} & \mathbf{95.8} & \mathbf{90.3} & 91.3 & 65.1\\
    \text{RaST} & 90.5 & 88.3 & 83.6 & 94.1 & 88.9 & 92.9 & 63.4\\
    \text{MST} & 92.9 & 90.1 & 84.2 & 94.7 & 88.9 & 93.1 & 62.2\\
    \text{HCT} & 92.7 & 90.2 & 85.1 & 94.4 & 89.3 & \mathbf{93.5} & \mathbf{65.3}\\
    \bottomrule
  \end{array}
  $$
  \caption{BLEU, ROUGE, and EM metrics on \rew{}.}
  \label{tab:rew}
\end{table}

\begin{figure}[t]
  \begin{tikzpicture}
    \begin{axis}[xlabel=\text{Lower bound per-rule phrase coverage (\%)},ylabel=\text{EM},
      ymin=23.6,ymax=25.1,
      legend pos=south west,width=0.495\textwidth,height=0.275\textwidth,
      cycle list/Dark2-3,every node near coord/.style={font=\small},
      width=\linewidth]
      \addplot+[mark=x,semithick,nodes near coords,
	visualization depends on={value \thisrow{val} \as \val},
	every node near coord/.append style={label={[yshift=-2.65em]\scriptsize$(\val)$}}]
	table {
	  x	y	val
	  0.225	24.5	43
	  0.5	24.9	24
	  0.75	24.0	19
	  1.1	23.8	13
	};
    \end{axis}
  \end{tikzpicture}
  \caption{Effect of rule clustering threshold parameter on EM for \cnr{}. Rule count is shown in parentheses.}
  \label{fig:rule-trend}
\end{figure}
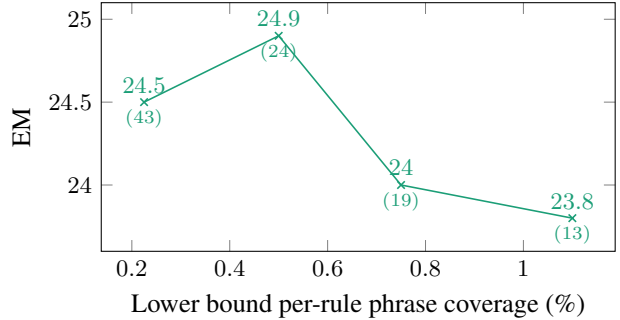

\subsection{Analysis}
Here we analyze how properties of rule vocabularies affect HCT's rewriting performance. We also examine a few generated rewrites to compare model behavior.

\begin{table}[t]
  \small
  \begin{tabular}{>{\scriptsize}r@{\hskip 0.625em}| X l >{\scriptsize}r@{\hskip 0.625em} | X l >{\scriptsize}r@{\hskip 0.625em} | X l}
    1 & \bla{} & 9 & \bla{} by \bla{} & 17 & for \bla{}\\
    2 & \bla{} 's & 10 & \bla{} than \bla{} & 18 & in \bla{}\\
    3 & \bla{} \bla{} & 11 & \bla{} than \bla{} \bla{} & 19 & in \bla{} \bla{} \bla{}\\
    4 & \bla{} \bla{} ' & 12 & aside from \bla{} & 20 & of \bla{}\\
    5 & \bla{} \bla{} 's & 13 & besides \bla{} & 21 & other than \bla{}\\
    6 & \bla{} \bla{} 's \bla{} & 14 & besides \bla{} \bla{} & 22 & other than \bla{} \bla{} \bla{}\\
    7 & \bla{} \bla{} \bla{} & 15 & besides \bla{} \bla{} \bla{} & 23 & the \bla{} \bla{}\\
    8 & \bla{} \bla{} \bla{} 's & 16 & by \bla{} & 24 & to \bla{}\\
  \end{tabular}
  \caption{HCT rule vocabulary for \cnr{}.}
  \label{tab:rules}
\end{table}
\begin{table}[t]
  \small
  \begin{tabular}{>{\scriptsize}X r@{\hskip 0.625em}| X l@{\hskip 4.5em}>{\scriptsize}r@{\hskip 0.625em} | X l@{\hskip 4.5em}>{\scriptsize}X r@{\hskip 0.625em} | X l}
    1 & \bla{} & 5 & \bla{} \bla{} & 9 & okay\\
    2 & \bla{} 's & 6 & \bla{} \bla{} 's & 10 & song\\
    3 & \bla{} , \bla{} & 7 & of \bla{} & 11 & the \bla{}\\
    4 & \bla{} , \bla{} , & 8 & of \bla{} \bla{} & 12 & the \bla{} \bla{}\\
  \end{tabular}
  \caption{For \mud{}.}
  \label{tab:mud-rules}
\end{table}

\subsubsection{Rule vocabulary} An ideal rule vocabulary will maximize phrase coverage while minimizing rule quantity. Recall that after clustering rules extracted from the data, we filter out rare rules that fall below a certain frequency threshold. Higher thresholds filter rules more aggressively; the points from left to right in Figure~\ref{fig:rule-trend} map to 43, 24, 19, and 13 rules. Based on the slight bump in EM between 43 and 24 rules, too many low-frequency rules can be harmful. Yet EM drops dramatically after reducing vocabulary size below 24. Although we did not fully tune it in experiments, our chosen threshold of 0.5\% looks reasonable.

Table~\ref{tab:rules} lists 24 extracted rules from \cnr{} for more insight into their properties. Generally, we notice the use of possessive form (e.g., ``'s''), prepositions, and determiners. The rules also show redundancy as multiple groups of rules have the same tokens but differing slot counts; we note that HCT's BERT-encoded rule embeddings help it generalize within groups. The rule vocabularies for \mud{} (Table~\ref{tab:mud-rules}) and \rew{} are comparable to \cnr{}'s. Comparing \cnr{} and \mud{} rules, we observe that former's cover more prepositions phrases such as ``besides'' and ``other than''. This trait reflects domain differences; \cnr{} dialogues are longer and require more ellipsis resolution, while those in \mud{} often involve short user requests to a system. However, there are shared rules between the two datasets (e.g., ``\bla{} 's'', ``of \bla{}'', and ``the \bla{}'') that show shared patterns in missing target phrases.

\subsubsection{Semantic role labeling evaluation} Since source and target strings in \mud{} and \rew{} often show high overlap, models may easily achieve high scores on n-gram matching metrics such as BLEU and ROUGE\@. We follow \citet{hao2021robust} by augmenting these metrics with evaluation based on semantic role labeling (SRL). As SRL can be seen as a shallow version of semantic parsing, semantically similar predictions to the target should share SRL annotations. We apply state-of-the-art SRL models \cite{che2021n-ltp,shi2019simple} to extract the predicate-argument structure from predictions and targets on \mud{} and \rew{}. We then compute precision, recall, and F1 scores of the model-predicted predicate-argument spans. Table~\ref{tab:srl-eval} shows that HCT achieves significantly higher F1 scores than RaST and MST\@. Interestingly, the simpler RaST model outperforms MST in terms of this span-based SRL evaluation.

\section{Related Work}
\subsubsection{Iterative refinement} Instead of generating text in a single pass, iterative approaches repeatedly edit an output sequence. This applies to post-editing in machine translation \cite{novak2016iterative,xia2017deliberation,grangier2018quickedit}, allowing humans to refine hypotheses with a machine's help. These approaches have encoder-decoder architectures with two decoders. The first decoder attends over the source string to output a hypothesis, while the second attends to both the source and hypothesis to refine the latter. In contrast, the second level of HCT depends only on the initial source token and predicted rule embeddings; it does not re-encode the source after inserting predicted rules.

\subsubsection{Retrieval-based editing} Another approach that constrains the search space of text generation is the retrieve-then-edit framework. A model is trained to retrieve a prototype output, then edit it to match the input context \cite{hashimoto2018retrieve,wu2019response}. The intent is to encourage diverse outputs that remain grammatical and coherent. While we have similar motivations, the preceding work edits prototypes using attention-based seq-to-seq models. By limiting prediction length to that of the source string, our rule tagger and span predictor greatly reduce search space in comparison.

\subsubsection{Coreference resolution} The utterance rewriting task often requires coreference resolution, which has a long history of machine learning approaches. Recent neural models have eliminated parsers or hand-engineered algorithms, instead scoring span or mention pairs using distributed representations \cite{clark2016improving,lee2017end}. Pretraining methods like SpanBERT \cite{joshi2020spanbert} refine such span representations. \citet{wu2020corefqa} extend SpanBERT to frame coreference resolution as query-based span prediction. We go beyond coreferences by resolving ellipses as well.

\begin{table}[t]
  \[
    \begin{array}{l *{6}c}
      \toprule
      & \multicolumn{3}{c}{\text{\mud{}}} & \multicolumn{3}{c}{\text{\rew{}}}\\
      \cmidrule(lr){2-4}\cmidrule(lr){5-7}
      & \text{P} & \text{R} & \text{F1} & \text{P} & \text{R} & \text{F1}\\
      \midrule
      \text{RaST} & 82.9 & 83.4 & 83.1 & \mathbf{83.8} & 76.6 & 80.0\\
      \text{MST} & 82.9 & 80.4 & 81.6 & 80.7 & 76.7 & 78.6\\
      \text{HCT} & \mathbf{85.1} & \mathbf{84.2} & \mathbf{84.7} & 83.6 & \mathbf{79.8} & \mathbf{81.7}\\
      \bottomrule
    \end{array}
  \]
  \caption{SRL evaluation on \mud{} and \rew{}.}
  \label{tab:srl-eval}
\end{table}

\section{Conclusion}
This work proposed a hierarchical context tagger for simple, high coverage utterance rewriting. We first demonstrated the benefit of a multi-span tagger that generates spans autoregressively at each source position. Next, we showed how HCT improves flexibility of MST's span predictor by first predicting slotted rules, then replacing a fixed number of slots with spans. While the two systems apply the same action tagger for source token editing, HCT conditions its span predictor on rules for more constrained phrase insertion. We also described automatic labeling methods using LCS and syntactic alignment, in addition to a clustering technique for rule extraction. Experiments revealed that HCT surpasses several state-of-the-art utterance rewriting systems by large margins. Furthermore, HCT generalizes better than simpler model variants lacking a rule tagger on unseen data domains.

HCT has applications to text editing tasks with a high overlap between source and target sequences. These tasks are plentiful and include sentence fusion, grammar correction, and prototype editing for machine translation. Leveraging syntactic structures to further restrict tagger and span predictor search space is an intriguing future direction.

\bibliography{aaai22}

\begin{thebibliography}{38}
\providecommand{\natexlab}[1]{#1}

\bibitem[{Bahdanau, Cho, and Bengio(2015)}]{bahdanau2015neural}
Bahdanau, D.; Cho, K.; and Bengio, Y. 2015.
\newblock Neural machine translation by jointly learning to align and
  translate.
\newblock In \emph{International {Conference} on {Learning} {Representations}}.

\bibitem[{Che et~al.(2021)Che, Feng, Qin, and Liu}]{che2021n-ltp}
Che, W.; Feng, Y.; Qin, L.; and Liu, T. 2021.
\newblock N-{LTP}: An open-source neural language technology platform for
  {C}hinese.
\newblock In \emph{Proceedings of the 2021 Conference on Empirical Methods in
  Natural Language Processing: System Demonstrations}, 42--49.

\bibitem[{Chen and Cherry(2014)}]{chen2014systematic}
Chen, B.; and Cherry, C. 2014.
\newblock A systematic comparison of smoothing techniques for sentence-level
  {BLEU}.
\newblock In \emph{Proceedings of the Ninth Workshop on Statistical Machine
  Translation}, 362--367.

\bibitem[{Chiang(2005)}]{chiang2005hierarchical}
Chiang, D. 2005.
\newblock A hierarchical phrase-based model for statistical machine
  translation.
\newblock In \emph{Proceedings of the 43rd Annual Meeting on Association for
  Computational Linguistics}, 263--270.

\bibitem[{Choi et~al.(2018)Choi, He, Iyyer, Yatskar, Yih, Choi, Liang, and
  Zettlemoyer}]{choi2018quac}
Choi, E.; He, H.; Iyyer, M.; Yatskar, M.; Yih, W.-t.; Choi, Y.; Liang, P.; and
  Zettlemoyer, L. 2018.
\newblock QuAC: Question Answering in Context.
\newblock In \emph{Proceedings of the 2018 Conference on Empirical Methods in
  Natural Language Processing}, 2174--2184.

\bibitem[{Clark and Manning(2016)}]{clark2016improving}
Clark, K.; and Manning, C.~D. 2016.
\newblock Improving coreference resolution by learning entity-level distributed
  representations.
\newblock In \emph{Proceedings of the 54th Annual Meeting of the Association
  for Computational Linguistics (Volume 1: Long Papers)}, 643--653.

\bibitem[{Devlin et~al.(2019)Devlin, Chang, Lee, and
  Toutanova}]{devlin2019bert}
Devlin, J.; Chang, M.-W.; Lee, K.; and Toutanova, K. 2019.
\newblock BERT: Pre-training of deep bidirectional Transformers for language
  understanding.
\newblock In \emph{Proceedings of the 2019 Conference of the North American
  Chapter of the Association for Computational Linguistics: Human Language
  Technologies, Volume 1 (Long and Short Papers)}, 4171--4186.

\bibitem[{Elgohary, Peskov, and Boyd-Graber(2019)}]{elgohary2019can}
Elgohary, A.; Peskov, D.; and Boyd-Graber, J. 2019.
\newblock Can you unpack that? Learning to rewrite questions-in-context.
\newblock In \emph{Proceedings of the 2019 Conference on Empirical Methods in
  Natural Language Processing and the 9th International Joint Conference on
  Natural Language Processing (EMNLP-IJCNLP)}, 5918--5924.

\bibitem[{Frey and Dueck(2007)}]{frey2007clustering}
Frey, B.~J.; and Dueck, D. 2007.
\newblock Clustering by passing messages between data points.
\newblock \emph{Science}, 315(5814): 972--976.

\bibitem[{Grangier and Auli(2018)}]{grangier2018quickedit}
Grangier, D.; and Auli, M. 2018.
\newblock QuickEdit: Editing text \& translations by crossing words out.
\newblock In \emph{Proceedings of the 2018 Conference of the North American
  Chapter of the Association for Computational Linguistics: Human Language
  Technologies, Volume 1 (Long Papers)}, 272--282.

\bibitem[{Gu et~al.(2016)Gu, Lu, Li, and Li}]{gu2016incorporating}
Gu, J.; Lu, Z.; Li, H.; and Li, V.~O. 2016.
\newblock Incorporating copying mechanism in sequence-to-sequence learning.
\newblock In \emph{Proceedings of the 54th Annual Meeting of the Association
  for Computational Linguistics (Volume 1: Long Papers)}, 1631--1640.

\bibitem[{Hao et~al.(2021)Hao, Song, Wang, Xu, Tu, and Yu}]{hao2021robust}
Hao, J.; Song, L.; Wang, L.; Xu, K.; Tu, Z.; and Yu, D. 2021.
\newblock Robust dialogue utterance rewriting as sequence tagging.
\newblock In \emph{Proceedings of the 2021 Conference on Empirical Methods in
  Natural Language Processing}, 4913--4924.

\bibitem[{Hashimoto et~al.(2018)Hashimoto, Guu, Oren, and
  Liang}]{hashimoto2018retrieve}
Hashimoto, T.~B.; Guu, K.; Oren, Y.; and Liang, P. 2018.
\newblock A retrieve-and-edit framework for predicting structured outputs.
\newblock In \emph{Proceedings of the 32nd Conference on Neural Information
  Processing Systems}.

\bibitem[{Huang et~al.(2021)Huang, Li, Zou, and Zhang}]{huang2021sarg}
Huang, M.; Li, F.; Zou, W.; and Zhang, W. 2021.
\newblock SARG: A novel semi autoregressive generator for multi-turn incomplete
  utterance restoration.
\newblock In \emph{Proceedings of the AAAI Conference on Artificial
  Intelligence}, 13055--13063.

\bibitem[{Joshi et~al.(2020)Joshi, Chen, Liu, Weld, Zettlemoyer, and
  Levy}]{joshi2020spanbert}
Joshi, M.; Chen, D.; Liu, Y.; Weld, D.~S.; Zettlemoyer, L.; and Levy, O. 2020.
\newblock Span{BERT}: Improving pre-training by representing and predicting
  spans.
\newblock \emph{Transactions of the Association for Computational Linguistics},
  8: 64--77.

\bibitem[{Kiegeland and Kreutzer(2021)}]{kiegeland2021revisiting}
Kiegeland, S.; and Kreutzer, J. 2021.
\newblock Revisiting the weaknesses of reinforcement learning for neural
  machine translation.
\newblock In \emph{Proceedings of the 2021 Conference of the North American
  Chapter of the Association for Computational Linguistics: Human Language
  Technologies}, 1673--1681.

\bibitem[{Kitaev, Cao, and Klein(2019)}]{kitaev2019multilingual}
Kitaev, N.; Cao, S.; and Klein, D. 2019.
\newblock Multilingual constituency parsing with self-attention and
  pre-training.
\newblock In \emph{Proceedings of the 57th Annual Meeting of the Association
  for Computational Linguistics}, 3499--3505.

\bibitem[{Kumar and Joshi(2017)}]{kumar2017incomplete}
Kumar, V.; and Joshi, S. 2017.
\newblock Incomplete follow-up question resolution using retrieval based
  sequence to sequence learning.
\newblock In \emph{Proceedings of the 40th International ACM SIGIR Conference
  on Research and Development in Information Retrieval}, 705--714.

\bibitem[{Lee et~al.(2017)Lee, He, Lewis, and Zettlemoyer}]{lee2017end}
Lee, K.; He, L.; Lewis, M.; and Zettlemoyer, L. 2017.
\newblock End-to-end neural coreference resolution.
\newblock In \emph{Proceedings of the 2017 Conference on Empirical Methods in
  Natural Language Processing}, 188--197.

\bibitem[{Li et~al.(2016)Li, Galley, Brockett, Gao, and
  Dolan}]{li2016diversity}
Li, J.; Galley, M.; Brockett, C.; Gao, J.; and Dolan, W.~B. 2016.
\newblock A diversity-promoting objective function for neural conversation
  models.
\newblock In \emph{Proceedings of the 2016 Conference of the North American
  Chapter of the Association for Computational Linguistics: Human Language
  Technologies}, 110--119.

\bibitem[{Lin and Hovy(2002)}]{lin2002manual}
Lin, C.-Y.; and Hovy, E. 2002.
\newblock Manual and automatic evaluation of summaries.
\newblock In \emph{Proceedings of the ACL-02 Workshop on Automatic
  Summarization}, 45--51.

\bibitem[{Liu et~al.(2020)Liu, Chen, Lou, Zhou, and Zhang}]{liu2020incomplete}
Liu, Q.; Chen, B.; Lou, J.-G.; Zhou, B.; and Zhang, D. 2020.
\newblock Incomplete utterance rewriting as semantic segmentation.
\newblock In \emph{Proceedings of the 2020 Conference on Empirical Methods in
  Natural Language Processing (EMNLP)}, 2846--2857.

\bibitem[{Manning et~al.(2014)Manning, Surdeanu, Bauer, Finkel, Bethard, and
  McClosky}]{manning2014stanford}
Manning, C.~D.; Surdeanu, M.; Bauer, J.; Finkel, J.~R.; Bethard, S.; and
  McClosky, D. 2014.
\newblock The Stanford {CoreNLP} natural language processing toolkit.
\newblock In \emph{Proceedings of 52nd Annual Meeting of the Association for
  Computational Linguistics: System Demonstrations}, 55--60.

\bibitem[{Novak, Auli, and Grangier(2016)}]{novak2016iterative}
Novak, R.; Auli, M.; and Grangier, D. 2016.
\newblock Iterative refinement for machine translation.
\newblock \emph{arXiv preprint arXiv:1610.06602}.

\bibitem[{Pan et~al.(2019)Pan, Bai, Wang, Zhou, and Liu}]{pan2019improving}
Pan, Z.; Bai, K.; Wang, Y.; Zhou, L.; and Liu, X. 2019.
\newblock Improving open-domain dialogue systems via multi-turn incomplete
  utterance restoration.
\newblock In \emph{Proceedings of the 2019 Conference on Empirical Methods in
  Natural Language Processing and the 9th International Joint Conference on
  Natural Language Processing (EMNLP-IJCNLP)}, 1824--1833.

\bibitem[{Papineni et~al.(2002)Papineni, Roukos, Ward, and
  Zhu}]{papineni2002bleu}
Papineni, K.; Roukos, S.; Ward, T.; and Zhu, W.-J. 2002.
\newblock {BLEU}: a method for automatic evaluation of machine translation.
\newblock In \emph{Proceedings of the 40th Annual Meeting of the Association
  for Computational Linguistics}, 311--318.

\bibitem[{Reddy, Chen, and Manning(2019)}]{reddy2019coqa}
Reddy, S.; Chen, D.; and Manning, C.~D. 2019.
\newblock CoQA: A conversational question answering challenge.
\newblock \emph{Transactions of the Association for Computational Linguistics},
  7: 249--266.

\bibitem[{See, Liu, and Manning(2017)}]{see2017get}
See, A.; Liu, P.~J.; and Manning, C.~D. 2017.
\newblock Get to the point: Summarization with pointer-generator networks.
\newblock In \emph{Proceedings of the 55th Annual Meeting of the Association
  for Computational Linguistics (Volume 1: Long Papers)}, 1073--1083.

\bibitem[{Shi and Lin(2019)}]{shi2019simple}
Shi, P.; and Lin, J. 2019.
\newblock Simple {BERT} models for relation extraction and semantic role
  labeling.
\newblock \emph{arXiv preprint arXiv:1904.05255}.

\bibitem[{Su et~al.(2019)Su, Shen, Zhang, Sun, Hu, Niu, and
  Zhou}]{su2019improving}
Su, H.; Shen, X.; Zhang, R.; Sun, F.; Hu, P.; Niu, C.; and Zhou, J. 2019.
\newblock Improving multi-turn dialogue modelling with utterance rewriter.
\newblock In \emph{Proceedings of the 57th Annual Meeting of the Association
  for Computational Linguistics}, 22--31.

\bibitem[{Tseng et~al.(2021)Tseng, Bhargava, Lu, Moniz, Piraviperumal, Li, and
  Yu}]{tseng2021cread}
Tseng, B.-H.; Bhargava, S.; Lu, J.; Moniz, J. R.~A.; Piraviperumal, D.; Li, L.;
  and Yu, H. 2021.
\newblock CREAD: Combined resolution of ellipses and anaphora in dialogues.
\newblock In \emph{Proceedings of the 2021 Conference of the North American
  Chapter of the Association for Computational Linguistics: Human Language
  Technologies}, 3390--3406.

\bibitem[{Tu et~al.(2016)Tu, Lu, Liu, Liu, and Li}]{tu2016modeling}
Tu, Z.; Lu, Z.; Liu, Y.; Liu, X.; and Li, H. 2016.
\newblock Modeling coverage for neural machine translation.
\newblock In \emph{Proceedings of the 54th Annual Meeting of the Association
  for Computational Linguistics (Volume 1: Long Papers)}, 76--85.

\bibitem[{Wolf et~al.(2020)Wolf, Debut, Sanh, Chaumond, Delangue, Moi, Cistac,
  Rault, Louf, Funtowicz, Davison, Shleifer, von Platen, Ma, Jernite, Plu, Xu,
  Scao, Gugger, Drame, Lhoest, and Rush}]{wolf2020transformers}
Wolf, T.; Debut, L.; Sanh, V.; Chaumond, J.; Delangue, C.; Moi, A.; Cistac, P.;
  Rault, T.; Louf, R.; Funtowicz, M.; Davison, J.; Shleifer, S.; von Platen,
  P.; Ma, C.; Jernite, Y.; Plu, J.; Xu, C.; Scao, T.~L.; Gugger, S.; Drame, M.;
  Lhoest, Q.; and Rush, A.~M. 2020.
\newblock Transformers: state-of-the-art natural language processing.
\newblock In \emph{Proceedings of the 2020 Conference on Empirical Methods in
  Natural Language Processing: System Demonstrations}, 38--45.

\bibitem[{Wu et~al.(2020)Wu, Wang, Yuan, Wu, and Li}]{wu2020corefqa}
Wu, W.; Wang, F.; Yuan, A.; Wu, F.; and Li, J. 2020.
\newblock CorefQA: Coreference resolution as query-based span prediction.
\newblock In \emph{Proceedings of the 58th Annual Meeting of the Association
  for Computational Linguistics}, 6953--6963.

\bibitem[{Wu et~al.(2019)Wu, Wei, Huang, Wang, Li, and Zhou}]{wu2019response}
Wu, Y.; Wei, F.; Huang, S.; Wang, Y.; Li, Z.; and Zhou, M. 2019.
\newblock Response generation by context-aware prototype editing.
\newblock In \emph{Proceedings of the AAAI Conference on Artificial
  Intelligence}, volume~33, 7281--7288.

\bibitem[{Xia et~al.(2017)Xia, Tian, Wu, Lin, Qin, Yu, and
  Liu}]{xia2017deliberation}
Xia, Y.; Tian, F.; Wu, L.; Lin, J.; Qin, T.; Yu, N.; and Liu, T.-Y. 2017.
\newblock Deliberation networks: Sequence generation beyond one-pass decoding.
\newblock \emph{Advances in Neural Information Processing Systems}, 30:
  1784--1794.

\bibitem[{Yu et~al.(2019)Yu, Zhang, Yasunaga, Tan, Lin, Li, Er, Li, Pang, Chen
  et~al.}]{yu2019sparc}
Yu, T.; Zhang, R.; Yasunaga, M.; Tan, Y.~C.; Lin, X.~V.; Li, S.; Er, H.; Li,
  I.; Pang, B.; Chen, T.; et~al. 2019.
\newblock SParC: Cross-Domain Semantic Parsing in Context.
\newblock In \emph{Proceedings of the 57th Annual Meeting of the Association
  for Computational Linguistics}, 4511--4523.

\bibitem[{Zhang et~al.(2019)Zhang, Ma, Duh, and Van~Durme}]{zhang2019amr}
Zhang, S.; Ma, X.; Duh, K.; and Van~Durme, B. 2019.
\newblock AMR parsing as sequence-to-graph transduction.
\newblock In \emph{Proceedings of the 57th Annual Meeting of the Association
  for Computational Linguistics}, 80--94.

\end{thebibliography}

\clearpage
\appendix
\section{Semi-Autoregressive Span Attention}
\label{sec:appendix}
To predict start index $s_{ij}^\uparrow$ in Eq.~\ref{eq:sp-sing}, we compute attention between source token $x_i$ and all context tokens semi-autoregressively. Instead of using the same source token embedding across spans to attend to context tokens, we use span-level embeddings $\mathbf{u}_{ij}$ (Eq.~\ref{eq:sp-attn1}), which depends on the below components.
\begin{enumerate}
  \item $\hat{\mathbf{u}}_{ij}$: A mixture of context token embeddings weighted by the previous span's attention distribution. Attention coefficients $\alpha_{i(j - 1)k}$ (Eq.~\ref{eq:sp-attn2}) over context tokens indexed by $k$ represent the attention distribution at span $j - 1$.
  \item $\mathbf{u}_{i(j - 1)}$: The previous span's embedding for source $x_i$.
\end{enumerate}
Note that at $j = 0$, the previous span's attention distribution is initialized as uniform over all context tokens. In addition, $\mathbf{u}_{i(j - 1)}$ is initialized to $\mathbf{e}_i$ from the source encoder.

\end{document}